\documentclass[11pt]{article}
\pdfoutput=1
\usepackage[margin=1in]{geometry}
\usepackage[T1]{fontenc}
\usepackage[utf8]{inputenc}
\usepackage{lmodern}
\usepackage{microtype}
\usepackage{graphicx,booktabs,amsmath,amssymb}
\usepackage[round]{natbib}
\usepackage{listings}
\usepackage[hidelinks]{hyperref}
\providecommand{\tightlist}{\setlength{\itemsep}{0pt}\setlength{\parskip}{0pt}}
\title{Retrieved but not ranked: surface-form bias in structural retrieval, from mathematics to agent trajectories}
\author{Nabira Rashid\textsuperscript{1} \and Manolis Kellis\textsuperscript{2}\\[6pt]
\small \textsuperscript{1}Independent; work conducted while contributing to Mantis at MIT CSAIL Kellis Lab \quad \textsuperscript{2}MIT CSAIL}
\date{August 2026}
\begin{document}
\maketitle

\begin{abstract}

Embedding retrieval is usually validated where surface form and meaning point the same way. We study the case where they are pulled apart on purpose, retrieving items that share underlying structure but not wording, in two unrelated domains under one protocol: competition mathematics (MathNet-Retrieve; 500 queries against a 117,088-item corpus) and embodied-agent trajectories (ALFWorld-derived; 118 queries against 336 trajectories). In mathematics the failure is complete and easy to locate. Strict Hit@1 at the heaviest disguise tier is 0.0\% for both production embedders (bootstrap 95\% CI {[}0.0, 0.0{]}), the correct item sits in the top 10 nearly always, and in 95.2 to 99.8\% of misses the candidate that wins is more lexically similar to the query than the correct answer. In trajectories, where surface variation is incidental rather than adversarial, the same models land at or near hypergeometric chance when gold must involve a different target object, and below chance for all three embedders once gold must involve a different object and receptacle: retrieval is anchoring on literal tokens rather than task structure. A lexical reranker control hurts in mathematics and helps in trajectories (closing 26 to 36\% of the recoverable gap, CIs excluding zero). Which sign it takes depends on how each benchmark's surface variation was built, adversarial or incidental, and that makes the control a cheap diagnostic. An LLM reranker recovers 5 to 63\% of the gap in mathematics and 43 to 76\% in trajectories. Direction replicates across three independently trained judges (all twenty-one judge-by-configuration cells positive), and nothing about magnitude transfers: effect sizes, tier profiles, and even which judge is the outlier all change with domain, with paired bootstrap differences between judges excluding zero in every configuration. Mathematics reranking gains concentrate on well-known competitions (+19.8 points, CI {[}+6.7, +33.2{]} in one of six judge-by-candidate cells), so part of the recovery, though not all of it, is memorization. Finally, in a paired downstream experiment (210 queries, two graders at 96 to 99\% agreement), oracle retrieval was statistically indistinguishable from adversarially bad retrieval (McNemar p = 0.678), and a complete-answers-only analysis explains why: the solver's 69.5\% zero-shot accuracy is largely a truncation proxy, with 97 to 100\% accuracy on answers that finish within budget, which leaves retrieval almost no headroom to act on.

\end{abstract}

\section{Introduction}\label{introduction}

Most retrieval benchmarks reward the easy case. A query about quadratic equations retrieves documents containing the words ``quadratic equation,'' wording and meaning move together, and embedding models trained against such benchmarks inherit that alignment. Structural retrieval breaks the alignment deliberately. Two mathematics problems can call for the same technique while sharing no vocabulary. Two agent trajectories can run the same procedure over completely different objects. Near-identical text can hide a flipped inequality, or a different household task. When surface and structure come apart, which one does embedding retrieval follow?

The question cannot be settled in one domain, because a single-domain answer mixes properties of embedding retrieval with properties of that benchmark's construction. So we run one protocol across two domains that have nothing in common, and this is what turns up the result we consider most useful: a cheap lexical control flips sign depending on how the benchmark was built. No single-domain study could see that, and it changes how ``semantic'' retrieval evaluations should be read.

Our contributions:

\begin{itemize}
\tightlist
\item
  \textbf{A two-domain evaluation of structural retrieval} under a shared protocol: tiered disguise-requiring gold, exact chance baselines, rank-1 failure taxonomy, and bootstrap confidence intervals on every headline number. Each source benchmark's published numbers were reproduced as a validation gate before we trusted the pipeline for anything else.
\item
  \textbf{Evidence that embedding retrieval anchors on literal surface content in both domains}: zero rank-1 accuracy under heavy disguise in mathematics with the answer retrievable below, and below-chance retrieval in trajectories once gold excludes the query's literal object and receptacle tokens.
\item
  \textbf{The lexical-control sign flip.} The same lexical reranker damages mathematical retrieval and improves trajectory retrieval. A token-class ablation locates the mechanism, and we suggest the control's sign as a cheap diagnostic of a benchmark's surface-variation regime.
\item
  \textbf{A judge-dependence result for LLM reranking}: recovery direction replicates across three independently trained judges in both domains, while effect size, tier profile, and even which judge is the outlier change with domain, and one judge's gains drop by more than a third across query styles within a domain. Neither a reranking effect size nor a judge ranking is portable.
\item
  \textbf{A contamination probe reported at its measured strength}: reranking gains concentrate on well-known competitions, confidently non-null in one of six judge-by-candidate cells under bootstrap resampling and directionally consistent elsewhere.
\item
  \textbf{A paired downstream null, with its mechanism located}: with a deliberately bad retrieval condition that prior work lacked, oracle and adversarial retrieval are indistinguishable, and a complete-answers-only analysis shows the headline 69.5\% zero-shot accuracy to be largely a truncation proxy, leaving near-zero effective headroom for retrieval to act on.
\item
  \textbf{Eight documented evaluation-integrity incidents}, most involving silent truncation that parsed as valid output.
\end{itemize}

\section{Related work}\label{related-work}

\textbf{Structural retrieval benchmarks.} MathNet \citep{alshammari2026mathnet} showed that 27 embedding models fail mathematical-equivalence retrieval (Recall@1 below 5\%) while recall stays high, and attributed the failure to superficial overlap. It did not test reranking. The procedural-memory benchmark of \citet{kohar2025benchmark} found the same kind of generalization cliff for trajectory retrieval on ALFWorld \citep{shridhar2021alfworld}, with three limitations we address: a single 384-dimensional encoder \citep{reimers2019sentence}, relevance labels from an LLM judge with modest human agreement (Cohen's kappa 0.178), and two-stage retrieval left as future work. We supply that stage on their released data with modern embedders, scored against exhaustive task-type labels instead of judged pools, which also avoids a pool-coverage artifact we document in their protocol.

\textbf{Retrieve-then-rerank} is standard information retrieval \citep{nogueira2019passage}. We contribute nothing to the architecture; what we contribute is its evaluation under structural relevance instead of topical relevance, with controls built for LLM judges.

\textbf{LLM-as-judge} scales evaluation at the price of importing the judge's training distribution \citep{zheng2023judging}. Our contamination analysis makes that concern concrete for reranking: where gains concentrate on items the judge plausibly memorized, claims about reasoning need discounting.

\textbf{Context that hurts.} Irrelevant context degrades LLM problem solving \citep{shi2023large}. MathNet observed embedding-retrieval RAG scoring below zero-shot for three solvers and did not investigate. We measure the link with a paired design and a deliberately bad condition.

\section{Shared protocol}\label{shared-protocol}

Both domains run through one pipeline. We specify it fully here so that every result in Sections 4 through 6 can be reconstructed from this section alone.

\subsection{Query and corpus construction}\label{query-and-corpus-construction}

\textbf{Mathematics.} Queries are 500 items from MathNet-Retrieve. Each query is a paraphrase of a source competition problem at one of two disguise tiers, EASY or HARD, written by the benchmark's authors with a Gemini-family model (a provenance fact we return to in Section 4). The corpus is the full 117,088-item MathNet collection, and we rank all of it for every query; nothing is shortlisted before retrieval. Each query has exactly one gold item, its source problem, and the benchmark plants a lexically similar but structurally different near-miss for every query.

\textbf{Agent trajectories.} The corpus is 336 ALFWorld trajectories. Queries are 118 trajectories: the 40 released by the source benchmark plus 78 we drew from ALFWorld's public split to raise statistical power. Each trajectory is a task instruction followed by an action sequence, and ALFWorld's six task types (for example \emph{pick and place}, \emph{heat and place}, \emph{examine in light}) define what counts as a procedure.

\subsection{Gold definition}\label{gold-definition}

\textbf{Mathematics} uses MathNet's expert labels unchanged: the gold item is the source problem. Strict Hit@k counts only the gold; lenient Hit@k also counts the planted near-miss. We report both because the gap between them is the paper's central measurement.

\textbf{Trajectories.} Relevance is defined by task type. We labeled all 336 corpus items exhaustively from ALFWorld's task metadata (rule-derived from the task metadata, then human-audited on stratified 60-item samples of both corpus and queries with zero disagreements), which removes the source benchmark's pooled-relevance problem. The rule classifier is a set of eleven anchored regular expressions over the corpus's 180 unique task-description strings, one template family per task type; every trajectory is tested against all eleven, and any cross-type collision is flagged as ambiguous rather than resolved by rule order. On top of that we apply two disguise requirements of increasing strictness: definition (i) requires a same-type gold to involve a different target object; definition (ii) requires a different object \emph{and} a different receptacle, so that gold and query share neither of the tokens most likely to drive lexical similarity. Because every candidate carries a label, chance is exact and hypergeometric for each query, and we report retrieval against it.

\subsection{Embedding retrieval}\label{embedding-retrieval}

We embed each query once and rank it against the full corpus by cosine similarity. In trajectories the corpus text is the task instruction followed by the full numbered state and action sequence, while the query text is the instruction alone; the LLM judges below see exactly these same texts. Mathematics uses two production embedders, gemini-embedding-001 and Qwen3-Embedding-8B (served via DeepInfra). Trajectories add MiniLM-L6-v2, the source benchmark's own encoder, kept as the weak-baseline contrast, and use the Qwen model served on a lab-hosted deployment. The deployment comparison in Section 4 embeds the same 500 texts through both Qwen servings and reruns retrieval under each.

\subsection{Reranking stages}\label{reranking-stages}

Reranking always operates on the same fixed top-10 candidate list from the embedding stage, so every condition is compared on identical inputs.

\textbf{Lexical control.} Before any LLM sees a candidate, a deliberately naive reranker re-sorts the top-10 by a lexical score that averages three signals, each in {[}0, 1{]}: Jaccard overlap of lowercased word tokens without stemming, a Levenshtein edit-distance ratio, and a length ratio. In mathematics it compares full problem texts; in trajectories it compares the query instruction with the candidate's task description. The mechanism ablation in Section 5 uses a narrower measure, plain Jaccard over restricted token sets: a closed list of fourteen ALFWorld action verbs for the verb-only variant, and the residual non-verb, non-stopword tokens for the noun-only variant (no part-of-speech tagger is involved). Its job is diagnostic. It can only add surface signal, so its sign tells us whether a benchmark's surface variation is adversarial or incidental.

\textbf{LLM judges.} Three independently trained judges rerank the same top-10: Gemini 3.1 Flash-Lite, GLM-5.2 (fp8, served on a lab endpoint), and Claude Haiku 4.5, all at temperature 0. We use two prompt variants, terse and chain-of-thought (verbatim prompts in Appendix~\ref{app:prompts}). Each judge returns a single best candidate, which is placed at rank 1 with the remaining embedding order retained, so reranked Hit@1 is computed on that pick; unparseable outputs fall back to the embedding order and are counted in the truncation audit.

\subsection{Metrics and statistics}\label{metrics-and-statistics}

The primary metric is \textbf{share of the recoverable gap closed}: (Hit@1 after reranking minus Hit@1 before) divided by (Hit@10 minus Hit@1 before). A reranker is therefore scored only against what its candidate list made retrievable. All Hit@k values carry bootstrap 95\% confidence intervals from resampling queries (10,000 resamples, seed 12345); share-of-gap intervals use the same resample count with seed 54321. Judge comparisons use \textbf{paired} bootstrap differences on the same queries (10,000 resamples, seed 42), reported for ten pre-specified pairs (four in mathematics, six in trajectories). Contamination effects use bootstrap intervals on the difference between famous-competition and other-source candidates. The deployment comparison uses McNemar's test on per-query hit flips, and the utility experiment uses McNemar's test on paired per-problem correctness.

\subsection{Downstream utility design}\label{downstream-utility-design}

A single solver, DeepSeek-v4-flash at a 32,768-token output budget, attempts 210 MathNet problems under three conditions: no context, a deliberately bad retrieval (a lexically similar but structurally unrelated problem), and gold-with-solution. Two independent graders that never include the solver mark the answers, Gemini 3.1 Flash-Lite (primary) and GLM-5.2 (robustness check); each returns an integer score from 0 to 7, binarized as correct at 6 or above, and they agree on 96 to 99\% of items (grader prompt in Appendix~\ref{app:grader}). We log finish-reason for every generation so that truncated outputs can be separated from completed ones.

\subsection{Validation gates}\label{validation-gates}

Before any new measurement, each domain reproduces a published number with the original artifacts. MathNet's Table 4 easy-tier values are recovered within one point (12.2, 89.8, and 97.6\% for Hit@1, Hit@5, and Hit@10 against the published 11.36, 90.68, and 96.93\%), and the trajectory benchmark's MEDIUM mean average precision is recovered within 0.007 (0.753 against the published 0.746) using the source paper's own encoder. A deviation beyond either tolerance would have halted the study.

\subsection{Integrity audits}\label{integrity-audits}

Every API response is cached and every reported number is recomputed from frozen aggregates. Standing audits include finish-reason checks on all generations, per-condition truncation rates, and spend reconciliation (final cost \$17.24 over 4,338 calls). Section 8 documents the eight incidents these audits surfaced.

\section{Domain 1: mathematical problems}\label{domain-1-mathematical-problems}

\textbf{Setup.} Section 3.1 gives the construction: 500 anchors against 117,088 items, each anchor with three LLM-generated equivalent reformulations (easy, medium, and hard disguise) and roughly three near-miss decoys that keep the surface form and change the mathematics. One design fact matters for everything below: the corpora are byte-identical across tiers and only the gold designation changes, so the tier axis isolates disguise and nothing else.

\textbf{Baseline.} Figure 1 and Table 1 report strict Hit@k. The hard-tier zero is degenerate, in the technical sense: no bootstrap resample of 500 queries ever contains a hit. Lenient Hit@10 is 99 to 100\% at both tiers. Retrieval succeeds; ranking fails.

\begin{figure}[t]
\centering
\includegraphics[width=\linewidth]{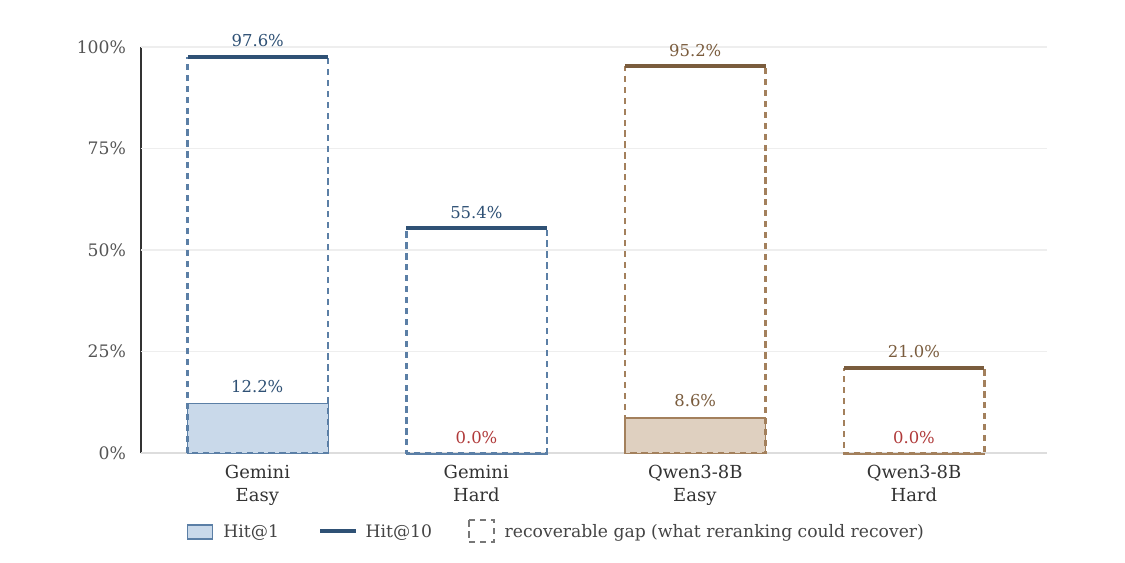}
\caption{The right answer is retrieved but not ranked first. Strict Hit@1 (bars) against Hit@10 (lines), 500 queries against the full 117,088-item corpus; the dashed span marks the recoverable gap available to any reranker.}
\label{fig:1}
\end{figure}

\begin{table}[t]
\centering
\small
\caption{Mathematics baseline, strict scoring, n=500. Brackets are bootstrap 95\% CIs.}
\label{tab:1}
\begin{tabular}{lllll}
\toprule
Embedder & Tier & Hit@1 & Hit@5 & Hit@10 \\
\midrule
Gemini-emb & Easy & 12.2\% [9.4, 15.2] & 89.8\% [87.0, 92.4] & 97.6\% [96.2, 98.8] \\
Gemini-emb & Hard & 0.0\% [0.0, 0.0] & 10.0\% [7.4, 12.8] & 55.4\% [51.0, 59.8] \\
Qwen-emb & Easy & 8.6\% [6.2, 11.0] & 86.8\% [83.8, 89.6] & 95.2\% [93.2, 97.0] \\
Qwen-emb & Hard & 0.0\% [0.0, 0.0] & 2.8\% [1.4, 4.4] & 21.0\% [17.6, 24.6] \\
\bottomrule
\end{tabular}
\end{table}

\textbf{Failure structure.} Between 84 and 98\% of misses, depending on embedder and tier, are the query's own planted near-miss. That is the trap working as designed, and we can measure how it works: in 95.2 to 99.8\% of misses the false positive is more lexically similar to the anchor than gold is, by the same lexical score the control uses. One representative case shows the mechanism. A character-identical problem with a single flipped inequality wins at cosine 0.860, while the true renamed-variable equivalent sits at rank four with cosine 0.821 (Appendix~\ref{app:ex-math}). The lexical reranker control reduces Hit@1 (a 9.1\% and 4.8\% share of gap lost at easy tier for Gemini-emb and Qwen-emb respectively; flat at hard tier). Surface similarity is anti-correlated with correctness here by construction, and there is no lexical generation fingerprint left for a reranker to exploit.

\textbf{LLM reranking.} All twelve terse judge-by-configuration cells are positive, and the judges' magnitudes separate beyond sampling noise (Figure 2 and Table 2; paired bootstrap differences between Gemini-j and GLM-j exclude zero in all four configurations). We had hoped a third judge would settle the ordering. It scrambled it instead. Haiku-j is 1.8 to 2.8 times stronger than Gemini-j and more than five times stronger than GLM-j at easy tier (58.1 and 63.3\% of gap closed), yet weaker than both at hard tier (5.4 and 6.7\%). It is the only judge whose gains degrade under heavier disguise the way naive intuition predicts; Gemini-j's hard-tier gains exceed its easy-tier gains, and GLM-j is flat to mildly rising across tiers. Three judges, three different tier profiles. Chain-of-thought doubles Gemini-j's easy-tier gains (44.7\% and 55.4\% share closed) while halving its hard-tier gains (22.7\% and 26.7\%). We expected the easy-tier improvement; the hard-tier drop surprised us. The traces offer one reading. The judge sometimes selects a candidate it itself calls ``a direct restatement,'' which suggests that room to deliberate lets it drift toward the most recognizable sibling. We consider this plausible and have not confirmed it. GLM-j chain-of-thought is excluded entirely: 63.3\% of its 2,000 responses truncated mid-reasoning while still parsing (Section 8).

\begin{figure}[t]
\centering
\includegraphics[width=\linewidth]{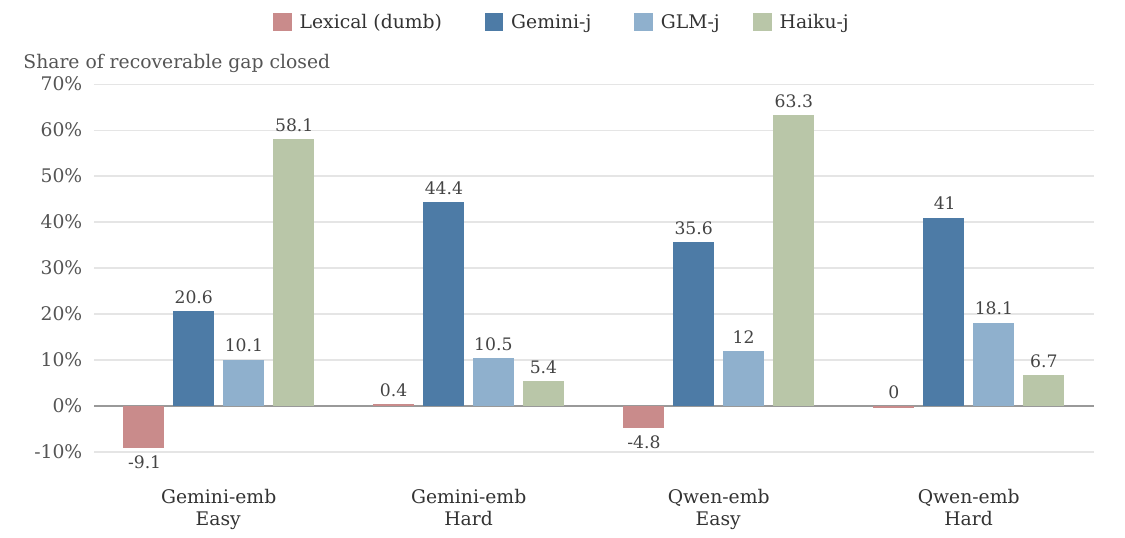}
\caption{Lexical reranking hurts while LLM reranking helps, mathematics domain. Share of the recoverable gap closed per configuration; all conditions rerank the same top-10 candidates.}
\label{fig:2}
\end{figure}

\begin{table}[t]
\centering
\small
\caption{Share of recoverable gap closed, terse prompt, mathematics. Brackets are bootstrap 95\% CIs.}
\label{tab:2}
\begin{tabular}{lllll}
\toprule
Candidates & Tier & Gemini-j & GLM-j & Haiku-j \\
\midrule
Gemini-emb & Easy & 20.6\% [14.7, 26.3] & 10.1\% [5.0, 15.0] & 58.1\% [52.7, 63.3] \\
Gemini-emb & Hard & 44.4\% [37.5, 51.3] & 10.5\% [6.9, 14.4] & 5.4\% [2.9, 8.3] \\
Qwen-emb & Easy & 35.6\% [29.9, 41.1] & 12.0\% [7.3, 16.6] & 63.3\% [58.2, 68.1] \\
Qwen-emb & Hard & 41.0\% [29.5, 53.3] & 18.1\% [10.5, 26.7] & 6.7\% [1.9, 12.4] \\
\bottomrule
\end{tabular}
\end{table}

\textbf{Contamination.} At hard tier, well-known competitions (IMO, USAMO, APMO; n=57) against the pooled rest (n=443) show the gaps in Table 3. One of six cells is confidently non-null. The other five, including both Haiku-j cells, point the same direction and are individually indistinguishable from zero at this sample size. Fisher tests on Gemini-j's four prompt-by-candidate combinations all reached p \textless{} 0.01, but we treat the bootstrap intervals as authoritative, so the summary is: contamination established in one cell, directionally consistent elsewhere, unproven as a judge-independent effect, and unmeasured under GLM-j, whose deliberative condition could not be scored. The chain-of-thought traces also show genuine technique-level reasoning, such as correctly identifying an invariant-mod-9 argument. Recognition and reasoning both operate, in proportions this design cannot separate.

\begin{table}[t]
\centering
\caption{Contamination gap (well-known minus rest), hard tier, terse prompt.}
\label{tab:3}
\begin{tabular}{llll}
\toprule
Judge & Candidates & Gap (pts) & 95\% CI \\
\midrule
Gemini-j & Gemini-emb & +19.8 & \textbf{[+6.7, +33.2]} \\
Gemini-j & Qwen-emb & +8.1 & [-1.4, +18.6] \\
GLM-j & Gemini-emb & +1.4 & [-5.0, +8.8] \\
GLM-j & Qwen-emb & +5.6 & [-1.2, +13.9] \\
Haiku-j & Gemini-emb & +4.5 & [-1.6, +12.0] \\
Haiku-j & Qwen-emb & +2.4 & [-1.6, +7.9] \\
\bottomrule
\end{tabular}
\end{table}

\textbf{Deployment divergence.} Two servings of identical Qwen3-Embedding-8B weights (mean pairwise cosine 0.9947 over 500 identical texts, never above 0.999, both unit-normalized) agree on five of six retrieval metrics. They differ significantly on the sixth, hard-tier Hit@10: 17 discordant queries in one direction against 1 in the other (McNemar exact p = 0.00014), and the difference lands exactly where gold-versus-decoy margins are thinnest. The practical lesson is that two deployments of the same model cannot be treated as interchangeable when ranking margins are thin.

\section{Domain 2: agent trajectories}\label{domain-2-agent-trajectories}

\textbf{Setup.} The released benchmark of \citet{kohar2025benchmark} supplies the 336 AgentInstruct ALFWorld trajectories and 40 of the queries; Section 3.1 describes the corpus and Section 3.2 the task-type gold. Three details belong here. The 78 added queries come from ALFWorld's public valid\_unseen split, every unique goal string available at a lightweight public source; the planned 150 was unreachable without a heavy simulation dependency, and we report the shortfall rather than patch it. The rule classifier's task-type labels were human-verified on stratified 60-item samples of both corpus and queries, with zero disagreements in both audits. And three benchmark corrections are documented in the repository: one released query relabeled on an unambiguous verb conflict (``chill'' implies cooling, not heating); the release's silently defaulting task-type field bypassed in favor of our own mapping; and the paper's 78-trajectory expert corpus, which is absent from the release, so all results use the AgentInstruct corpus.

\textbf{Tiers.} STRICT requires the same task type and a different target object (the disguise requirement). A harsher variant (ii) additionally requires a different receptacle. LENIENT accepts any same-task-type trajectory. Chance is computed per query via the exact hypergeometric tail, since gold-set sizes vary (mean strict set 51.2 of 336).

\textbf{Retrieval anchors on literal tokens.} Pooled strict Hit@1 sits at 17.8\% {[}11.0, 25.4{]} (Qwen-emb), 15.3\% {[}9.3, 22.0{]} (Gemini-emb), and 9.3\% {[}4.2, 14.4{]} (MiniLM) against chance of 15.3\%. The production models are at or barely above chance with CIs straddling the chance line; the weak encoder is below it. On its own that is ambiguous, and the definition-robustness check in Table 4 resolves it. Under definition (ii), the cleanest test of generalizing past the query's literal object and container tokens, all three embedders fall below chance (Figure 3). A model indifferent to structure would score at chance. Scoring systematically below it means the ranking is actively steered away from cross-object, cross-receptacle structural matches, and the thing doing the steering is token anchoring. The mechanism differs from the mathematics case (below chance here, zero-but-retrievable there); the direction of the bias is the same. Rankings follow the query's literal tokens.

\begin{figure}[t]
\centering
\includegraphics[width=\linewidth]{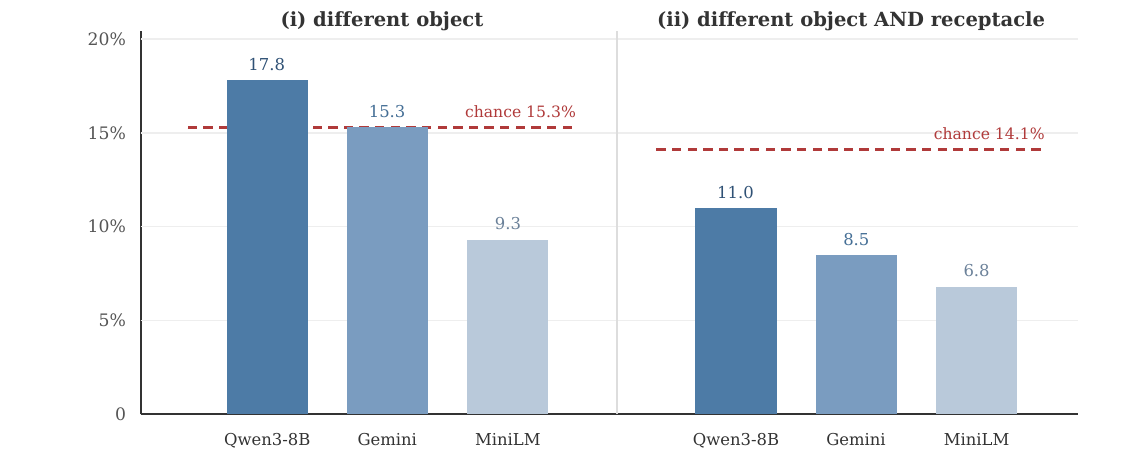}
\caption{Trajectory retrieval anchors on literal tokens. Strict Hit@1 for the three embedders against exact hypergeometric chance (dashed line), n=118, under gold definition (i), different object, and definition (ii), different object and receptacle. Under (ii) all three embedders fall below chance: sub-random, not indifferent.}
\label{fig:3}
\end{figure}

\begin{table}[t]
\centering
\caption{Strict Hit@1 versus exact chance under three gold definitions, n=118.}
\label{tab:4}
\begin{tabular}{lllll}
\toprule
Definition & Chance & Qwen-emb & Gemini-emb & MiniLM \\
\midrule
(i) different object & 15.3\% & 17.8\% & 15.3\% & 9.3\% \\
(ii) different object and receptacle & 14.1\% & \textbf{11.0\%} & \textbf{8.5\%} & \textbf{6.8\%} \\
(iii) any object (reference) & 16.9\% & 75.4\% & 74.6\% & 50.0\% \\
\bottomrule
\end{tabular}
\end{table}

We report the two query subsets separately because they behave differently. The original coverage-balanced 40 queries are substantially harder (strict Hit@1 10.0\% versus 21.8\% for Qwen-emb; 5.0\% versus 20.5\% for Gemini-emb) than the raw validation-split 78. This is a benchmark-construction effect confounded with phrasing style, and we flag it as such.

\textbf{The lexical control flips sign.} We ran this control expecting it to hurt, as it had in mathematics. It helped, and by enough to make it the trajectory domain's statistically firmest result: share of strict gap closed is +25.9\% {[}11.3, 41.2{]} (Qwen-emb), +36.4\% {[}23.4, 50.0{]} (Gemini-emb), and +32.1\% {[}18.5, 47.1{]} (MiniLM), all CIs excluding zero. A verb-only variant, run on the original-40 subset, reproduces most of the effect there, while a noun-only variant contributes little (0 to 12.5\% share closed). A within-domain phrasing slice (verb-first ``heat some X'' versus adjective-first ``put a hot X'', task types 3 to 5) shows the lexical reranker helping in every cell for both phrasings. The apparent contradiction between domains dissolves in the construction of the two benchmarks. MathNet's equivalents are adversarially paraphrased to suppress lexical overlap while its decoys preserve it, so lexical signal points exactly the wrong way. ALFWorld's surface variation is incidental, so verb and receptacle overlap is a real correlate of task type. The control costs nothing to compute, and its sign indicates which regime a benchmark occupies before any expensive evaluation runs.

\textbf{LLM reranking, and a judge reversal.} At n=118, terse judges close substantial strict gap for every embedder, and the judge ranking inverts relative to mathematics (Table 5). We had expected Gemini-j to lead here as it does there. Instead GLM-j, which trails Gemini-j by a factor of 2 to 4 in mathematics, leads by 1.3 to 1.7 times; paired bootstrap differences (GLM-j minus Gemini-j and GLM-j minus Haiku-j, same queries and candidate lists, 10,000 resamples) exclude zero for every embedder, including the one pairing where the marginal intervals overlap. The third judge settles which side is anomalous. Haiku-j lands within a few points of Gemini-j on the two production embedders and well inside its intervals on all three (43.9 to 48.2\% of gap closed), so two independent judges converge on this domain's magnitude while GLM-j alone diverges upward. Each domain therefore has a different outlier judge: Haiku-j in mathematics, GLM-j here. Provenance sharpens the picture. On the two production embedders' candidate sets, Gemini-j's gains drop steeply on the templated new-78 queries (36.6 to 40.0\%, versus 61.5 to 62.5\% on the human-paraphrased original 40), while its MiniLM cell is stable (56.3 versus 60.0) and GLM-j holds steady everywhere (68.3 to 76.0\%). Judge performance is sensitive to query style within a domain, and to domain itself. The judge-level failure taxonomy is consistent across all nine judge-by-embedder cells: siblings dominate (72 to 88\% of strict misses, near-misses 8 to 20\%), and the third judge replicates the pattern (per-cell tables in Appendix~\ref{app:taxonomy}). The judges have essentially solved the procedural confusion and fail almost exclusively on the disguise requirement, a categorically more benign error than the embedders' structural confusions. (Truncation audits: 1.8\% capped or unparsed in the two-judge run, isolated to GLM-j's known terse-output profile; 0.9\% in the Haiku-j run; both within threshold.)

\begin{table}[t]
\centering
\small
\caption{Share of recoverable gap closed, terse prompt, trajectories, n=118. Brackets are bootstrap 95\% CIs.}
\label{tab:5}
\begin{tabular}{llll}
\toprule
Candidates & Gemini-j & GLM-j & Haiku-j \\
\midrule
Qwen-emb & 42.6\% [27.3, 59.3] & \textbf{68.5\%} [50.9, 87.5] & 46.3\% [30.2, 63.0] \\
Gemini-emb & 45.5\% [31.9, 59.7] & \textbf{75.8\%} [60.7, 91.4] & 43.9\% [30.4, 57.9] \\
MiniLM & 58.9\% [42.1, 76.8] & \textbf{75.0\%} [56.9, 93.2] & 48.2\% [32.2, 64.8] \\
\bottomrule
\end{tabular}
\end{table}

\section{Downstream: does retrieval quality reach the solver?}\label{downstream-does-retrieval-quality-reach-the-solver}

MathNet reported gains of up to 12 points from expert retrieval and, without investigating, observed embedding-retrieval RAG scoring below zero-shot for three solvers. We measure the link with the condition their design lacked, deliberately bad retrieval, using the setup in Section 3.6.

Setup details that matter for reading the results: 210 mathematics queries; three conditions (no retrieval, adversarially reranked top-1, and the gold equivalent with its solution); solver DeepSeek-v4-flash at a 32,768-token budget; two independent graders at 96.2 to 98.6\% per-condition binary agreement, with the solver's developer excluded from grading. Truncation is 30.5 to 31.4\%, uniform across conditions. The condition-correlated truncation that invalidated an earlier pilot (Section 8) is absent here, though the residual rate never resolved across three cap increases (80\%, then 50\%, then 31\%) and is carried as a stated caveat on every accuracy below.

The result is a paired null, and the paired analysis sharpens it. Accuracy is flat (67 to 70\% in every condition, both graders). McNemar on none versus gold: 13 queries where context hurt against 10 where it helped (p = 0.678). None versus dumb: 11 against 8 (p = 0.648). Both graders concur. On the only subset where context could have acted, the 64 of 210 queries that failed zero-shot, gold recovers 10 (15.6\%) while breaking 13 of the 146 already solved, a net of minus 3. Deliberately bad retrieval nets the identical minus 3. For this solver, oracle retrieval was indistinguishable from adversarial retrieval.

A complete-answers-only check then locates the mechanism. Restricting to the 127 of 210 queries whose answers finished within budget in all three conditions, accuracy is 97.6 to 100\% in every condition under both graders, with zero to two discordant pairs per comparison (McNemar p between 0.50 and 1.00). In the no-context condition, completion and correctness nearly coincide under Grader A: the 146 finished answers and the 146 correct answers are both size 146 but overlap on 144, with two truncated answers credited for stating the right value before the cutoff and two finished derivations simply wrong; 206 of 210 queries align in total, and the equality of the two counts is a numerical coincidence rather than a set identity. The headline 69.5\% zero-shot accuracy is therefore closer to a proxy for whether the derivation fits the budget than to a measure of solving ability. The null holds for a specific reason: this solver has essentially no headroom on problems it can finish, so there was almost nothing for oracle context to improve. We report both framings because they are different claims. One says retrieval did not move accuracy; the other says there was almost nothing for it to move. A composition check addresses the natural follow-up, whether the ceiling reflects recognition of famous problems. 30 of the 210 queries come from well-known competitions, and near-ceiling accuracy on completable problems is fame-uniform (100.0\% versus 99.1\% under Grader A on the complete subset), so the ceiling does not reduce to fame. The well-known subset's higher raw accuracy in the full sample (80.0\% versus 67.8\%) tracks its lower truncation rate (26.7\% versus 31.1\%), which is what the completion-proxy reading predicts. Group counts are small and these figures are reported descriptively, without significance testing. This is one solver in one domain and we claim nothing wider. It remains a measured counterweight to the assumption that retrieval quality transfers downstream, with headroom now identified as the binding constraint rather than a conjectured moderator.

\section{Cross-domain synthesis}\label{cross-domain-synthesis}

\begin{enumerate}
\def\labelenumi{\arabic{enumi}.}
\item
  \textbf{Surface over structure, whenever the two are separable.} Mathematics shows 0\% strict rank-1 under heavy disguise with the answer retrievable below; trajectories show below-chance retrieval for all three embedders once gold excludes both the literal object and receptacle. The mechanisms differ. In both cases these embedding models rank by literal surface content.
\item
  \textbf{The lexical control's sign diagnoses the benchmark rather than the retriever.} Adversarial surface variation makes lexical signal anti-informative, incidental variation makes it informative, and a single-benchmark claim about ``semantic'' retrieval that omits this axis conflates the two regimes. The control costs nothing to run.
\item
  \textbf{LLM reranking recovers substantial gap in both regimes; direction is the only portable finding.} Across three judges, effect sizes span more than eightfold within a single configuration, tier profiles disagree (one judge improves under heavier disguise, one is flat, one degrades), each domain has a different outlier judge (paired bootstrap differences excluding zero in every configuration), and one judge's gains drop by more than a third across query styles within a domain. Any single reranking effect size is a property of the judge, prompt, domain, and query style jointly. Where provenance varies, memorization measurably contributes.
\item
  \textbf{Stages fail differently.} Embedder errors are structural confusions; judge errors are disguise failures over the correct structure (siblings, 72 to 88\% of judge misses in every trajectory cell). Pipeline evaluation should ask what kind of error survives each stage, and only then how many.
\item
  \textbf{The retrieval-to-solver link cannot be assumed.} Under a paired design with a bad-retrieval control, one competent solver extracted nothing from oracle context. A complete-answers-only analysis attributes this to near-zero effective headroom rather than to context being ignored, which is itself a caution about reading zero-shot accuracy under token budgets.
\end{enumerate}

\section{Evaluation-integrity findings}\label{evaluation-integrity-findings}

We record eight incidents. Most were caught by routine audits that existed only because of earlier incidents, and in every case we excluded the affected numbers rather than attempting repairs. (1) GLM-j chain-of-thought judging truncated at its token cap in 63.3\% of 2,000 responses while parsing successfully; a parsed response is not a concluded one, and the condition was excluded entirely. (2) A six-condition solver pilot truncated 49.6\% of answers, correlated with experimental condition (50 to 53\% with context versus 38.6\% without), which invalidated it outright; the oracle condition's accuracy on complete answers was 82.9\% against a reported 63.6\%. (3) A hidden-thinking-token quirk in one judge API silently burned output budgets twice before being fixed at the configuration level. (4) A spend-tracking discrepancy against the maintainer's own dashboard triggered a full stop and a call-by-call reconstruction from caches, closing the gap without new spend and producing code-enforced spend tracking. (5) Successive solver-cap increases reduced truncation from 80\% to 50\% to 31\% without resolving it, which suggests truncation tracks problem difficulty at least as much as budget. (6) A third judge initially appeared infeasible: DeepSeek-v4-flash as a reranker judge was still mid-reasoning with empty output at 24,000 tokens and 227 seconds on a single query, matching GLM-j's chain-of-thought behavior across a second provider, while a non-reasoning judge (Haiku-j) later completed the full 2,354-call run with 0.1 to 0.9\% truncation. What governs visible deliberation appears to be task framing rather than raw capability. (7) A shared-infrastructure solver run degraded fourfold under sustained load and was abandoned rather than waited out. (8) Bootstrap resampling tightened this project's own earlier contamination claim from ``positive in most cells'' to ``confidently non-null in one.'' Three further problems were caught before submission against frozen aggregates or raw caches: a suffix-matching scoring bug in the CI pass, a silent dependency drift that broke an embedder, and a set-equality claim built on two counts that coincided at 146 without being the same set. Standing practice: finish-reason audits after every generation run; exact-tokenizer verification of suspicious length distributions; per-condition truncation reporting wherever a budget cap exists; recomputation verified against frozen aggregates.

\section{Limitations}\label{limitations}

Mathematics uses a single fixed seed; bootstrap CIs quantify within-sample uncertainty only. Three judges is a small sample of judges. Magnitudes vary by more than eightfold among them, each domain has a different outlier, and tier profiles disagree, so the observed spread is best read as a lower bound on judge variability; one further candidate proved structurally infeasible (Section 8). The trajectory domain has one dataset family, one 336-item corpus, and 118 queries (short of the 150 target for stated availability reasons). Provenance and phrasing style are confounded in the old-versus-new subset contrast, and one judge's sensitivity to that contrast means judge results should not be assumed stable across query styles. Task-type labels are rule-inferred and human-verified on samples rather than authored. Contamination attribution is correlational; the well-known subset may differ in ways beyond training exposure. MathNet's equivalents and decoys were generated and partly verified by Gemini-3-flash, a same-family model as Gemini-j; family-level affinity to Gemini-generated paraphrases is a possible unmeasured contributor to Gemini-j's mathematics advantage and to the one significant contamination cell, which pairs a Gemini judge with Gemini-embedding candidates. The downstream null is one solver whose effective headroom proved near zero once truncation was controlled for; the 31\% solver-truncation rate is carried as a caveat and, per Section 6, is itself the binding constraint. Mathematics variants and decoys are LLM-generated; lexical controls constrain generation-process confounds without eliminating them.

\section{What this paper does not claim}\label{what-this-paper-does-not-claim}

That embedding retrieval is generally poor: lenient and easy-tier performance are strong in both domains. That lexical reranking is generally good or bad: its sign depends on benchmark construction, which is exactly what makes it informative. That LLM reranking solves structural retrieval: the majority of hard-tier gap remains unclosed. That memorization explains reranking gains: both mechanisms demonstrably operate. That downstream irrelevance generalizes beyond the measured solver. That the two domains share a mechanism: below-chance and zero-but-retrievable are different failure shapes, and the common claim is behavioral rather than mechanistic.

\section{Conclusion}\label{conclusion}

Across two unlike domains, embedding retrieval tracks literal surface content over underlying structure whenever the two are separable, to the point of ranking below random selection when gold requires generalizing past the query's literal tokens. A reranking stage recovers much of what ranking loses. Nothing about its magnitude transfers across judges, prompts, domains, or query styles, and part of the recovery is attributable to memorization where provenance varies. The most useful instrument in the study also turned out to be the cheapest: a lexical reranker whose sign reveals whether a benchmark's surface variation is adversarial or incidental. We suggest structural-retrieval evaluations report that control routinely, along with exact chance baselines and per-condition truncation audits.

\section{Reproducibility}\label{reproducibility}

Seed 42 in mathematics; all 118 available trajectory queries used, so no query sampling. All raw results, per-query cached judge and grader responses, spend logs, correction banners, and both benchmark-correction patches are in the repository; \texttt{results/FINAL\_NUMBERS.md} is the authoritative flat digest of every number cited here with its source file, and \texttt{results/RESULTS\_SUMMARY.md} maps each section to its script and raw output. Total experiment cost was \$17.24 across 4,338 recorded API calls. The repository is available at \url{https://github.com/nabirarashid/structural-retrieval}.

\section*{Acknowledgments}

Thanks to Anas Maarouf for access to the lab-hosted embedding and language-model endpoints used in parts of this study and for feedback on the draft, to Pranava Kumar for early discussions of the team's evaluation priorities that helped shape the direction, and to the Agents and Reasoning team at the MIT CSAIL Kellis Lab for mentorship and compute infrastructure.

\bibliographystyle{plainnat}
\bibliography{references}

\begin{thebibliography}{7}
\providecommand{\natexlab}[1]{#1}
\providecommand{\url}[1]{\texttt{#1}}
\expandafter\ifx\csname urlstyle\endcsname\relax
  \providecommand{\doi}[1]{doi: #1}\else
  \providecommand{\doi}{doi: \begingroup \urlstyle{rm}\Url}\fi

\bibitem[Alshammari et~al.(2026)Alshammari, Wen, Zainal, Hamilton, Safaei,
  Albarakati, Freeman, and Torralba]{alshammari2026mathnet}
S.~Alshammari, K.~Wen, A.~Zainal, M.~Hamilton, N.~Safaei, S.~Albarakati, W.~T.
  Freeman, and A.~Torralba.
\newblock {{MathNet}: A Global Multimodal Benchmark for Mathematical Reasoning
  and Retrieval}.
\newblock In \emph{ICLR 2026}, 2026.
\newblock arXiv:2604.18584.

\bibitem[Kohar and Krishnan(2025)]{kohar2025benchmark}
I.~Kohar and A.~Krishnan.
\newblock {A Benchmark for Procedural Memory Retrieval in Language Agents}.
\newblock \emph{arXiv preprint arXiv:2511.21730}, 2025.

\bibitem[Nogueira and Cho(2019)]{nogueira2019passage}
R.~Nogueira and K.~Cho.
\newblock {Passage Re-ranking with {BERT}}.
\newblock \emph{arXiv preprint arXiv:1901.04085}, 2019.

\bibitem[Reimers and Gurevych(2019)]{reimers2019sentence}
N.~Reimers and I.~Gurevych.
\newblock {{Sentence-BERT}: Sentence Embeddings using Siamese {BERT}-Networks}.
\newblock In \emph{EMNLP-IJCNLP 2019}, 2019.
\newblock arXiv:1908.10084.

\bibitem[Shi et~al.(2023)Shi, Chen, Misra, Scales, Dohan, Chi, Sch{\"a}rli, and
  Zhou]{shi2023large}
F.~Shi, X.~Chen, K.~Misra, N.~Scales, D.~Dohan, E.~Chi, N.~Sch{\"a}rli, and
  D.~Zhou.
\newblock {Large Language Models Can Be Easily Distracted by Irrelevant
  Context}.
\newblock In \emph{ICML 2023}, 2023.
\newblock arXiv:2302.00093.

\bibitem[Shridhar et~al.(2021)Shridhar, Yuan, C{\^o}t{\'e}, Bisk, Trischler,
  and Hausknecht]{shridhar2021alfworld}
M.~Shridhar, X.~Yuan, M.-A. C{\^o}t{\'e}, Y.~Bisk, A.~Trischler, and
  M.~Hausknecht.
\newblock {{ALFWorld}: Aligning Text and Embodied Environments for Interactive
  Learning}.
\newblock In \emph{ICLR 2021}, 2021.
\newblock arXiv:2010.03768.

\bibitem[Zheng et~al.(2023)Zheng, Chiang, Sheng, et~al.]{zheng2023judging}
L.~Zheng, W.-L. Chiang, Y.~Sheng, et~al.
\newblock {Judging {LLM}-as-a-Judge with {MT-Bench} and Chatbot Arena}.
\newblock In \emph{NeurIPS 2023}, 2023.
\newblock arXiv:2306.05685.

\end{thebibliography}

\appendix
\numberwithin{table}{section}

\section{Verbatim judge prompts}
\label{app:prompts}

All three judges (Gemini-j, GLM-j, Haiku-j) use identical prompt wording within each domain; the third judge's script imports the same template object the other two use rather than reimplementing it, so there is one terse prompt per domain rather than three. The text below is the exact string sent to the API. \texttt{\{anchor\}} and \texttt{\{candidates\}} are substituted with the anchor problem or task text and the ten numbered candidate texts before sending.

\subsection{Mathematics domain, terse prompt (Gemini-j, GLM-j, Haiku-j)}
\label{app:math-terse}

\begin{lstlisting}
You are given a math competition problem ("ANCHOR") and 10 candidate problems, labeled 1 through 10. Exactly one candidate relies on the same underlying mathematical technique or method as the anchor -- the same core idea you would actually use to solve it -- even though it may look completely different on the surface.

IGNORE surface-level similarity when deciding: shared variable names, shared wording, shared language, shared story framing (e.g. both about chessboards, or both in the same language), or shared numbers do NOT mean same technique. A candidate can look nearly identical in phrasing to the anchor and still use a completely different technique, and a candidate can look nothing like the anchor on the surface and still use the exact same technique.

Focus only on: what mathematical concept, theorem, or method would you actually use to solve each problem. Which candidate needs the same one as the anchor?

ANCHOR:
{anchor}

CANDIDATES:
{candidates}

Respond with ONLY the candidate number (1-10). No explanation, no other text.
\end{lstlisting}

\subsection{Mathematics domain, chain-of-thought prompt (Gemini-j and GLM-j only)}
\label{app:math-cot}

Same task framing as Appendix~\ref{app:math-terse}, with a step-by-step reasoning instruction and a structured final-answer marker.

\begin{lstlisting}
You are given a math competition problem ("ANCHOR") and 10 candidate problems, labeled 1 through 10. Exactly one candidate relies on the same underlying mathematical technique or method as the anchor -- the same core idea you would actually use to solve it -- even though it may look completely different on the surface.

IGNORE surface-level similarity when deciding: shared variable names, shared wording, shared language, shared story framing (e.g. both about chessboards, or both in the same language), or shared numbers do NOT mean same technique. A candidate can look nearly identical in phrasing to the anchor and still use a completely different technique, and a candidate can look nothing like the anchor on the surface and still use the exact same technique.

Focus only on: what mathematical concept, theorem, or method would you actually use to solve each problem. Which candidate needs the same one as the anchor?

ANCHOR:
{anchor}

CANDIDATES:
{candidates}

Think step by step. First state the core technique needed to solve the ANCHOR. Then, for each candidate in turn, briefly note the technique it actually needs and whether it matches the anchor's. After going through all 10, conclude.

End your response with your final answer on its own line, in exactly this format:
FINAL ANSWER: <candidate number>
\end{lstlisting}

\subsection{Mathematics domain, concise chain-of-thought variant (diagnostic only)}
\label{app:math-cot-concise}

This variant was tested on a 30-query sample after the prompt in Appendix~\ref{app:math-cot} caused 63.3\% of GLM-j's full chain-of-thought run to truncate mid-reasoning (Section 8, incident 1). It never replaced that prompt in a full run and is not one of the headline judge configurations; it is included because it is part of the truncation story.

\begin{lstlisting}
You are given a math competition problem ("ANCHOR") and 10 candidate problems, labeled 1 through 10. Exactly one candidate relies on the same underlying mathematical technique or method as the anchor -- the same core idea you would actually use to solve it -- even though it may look completely different on the surface.

IGNORE surface-level similarity when deciding: shared variable names, shared wording, shared language, shared story framing (e.g. both about chessboards, or both in the same language), or shared numbers do NOT mean same technique. A candidate can look nearly identical in phrasing to the anchor and still use a completely different technique, and a candidate can look nothing like the anchor on the surface and still use the exact same technique.

Focus only on: what mathematical concept, theorem, or method would you actually use to solve each problem. Which candidate needs the same one as the anchor?

ANCHOR:
{anchor}

CANDIDATES:
{candidates}

Think step by step, but be CONCISE -- this is a triage judgment, not a full solution write-up. State the core technique needed for the ANCHOR in one sentence. Then for each candidate, in one short phrase each, name its technique and say match or no match -- do not re-derive or fully solve any candidate. If you are torn between two candidates, pick the better one directly rather than re-litigating the comparison at length.

End your response with your final answer on its own line, in exactly this format:
FINAL ANSWER: <candidate number>
\end{lstlisting}

\subsection{Trajectory domain, terse prompt (Gemini-j, GLM-j, Haiku-j)}
\label{app:traj-terse}

No chain-of-thought variant exists for this domain; chain-of-thought was run only in mathematics.

\begin{lstlisting}
You are given an agent task instruction ("ANCHOR") and 10 candidate task trajectories, labeled 1 through 10. Exactly one candidate follows the same underlying PROCEDURE as the anchor -- the same transformation type (e.g. simple placement, clean-then-place, heat-then-place, cool-then-place, examine-with-light, two-object placement) -- even though it may involve a completely different object.

IGNORE which specific object is involved when deciding: matching object names, matching receptacles, or similar surface phrasing do NOT mean same procedure. A candidate can involve the exact same object as the anchor and still follow a different procedure, and a candidate can involve a completely different object and still follow the exact same procedure.

Focus only on: what sequence of actions / transformation type would you need to perform for each. Which candidate needs the same one as the anchor?

ANCHOR:
{anchor}

CANDIDATES:
{candidates}

Respond with ONLY the candidate number (1-10). No explanation, no other text.
\end{lstlisting}

\section{Worked examples}
\label{app:examples}

\subsection{Mathematics: the flipped-inequality near-miss}
\label{app:ex-math}

This is the representative case cited in Section 4. Query \texttt{sau\_2012\_bb6608}:

\begin{quote}
Determine all positive integers $n$ such that the inequality
\[\sqrt{x-1} + \sqrt{x-2} + \cdots + \sqrt{x-n} < x\]
holds for every real number $x \ge n$.
\end{quote}

Gold target, the easy-tier equivalent reformulation:

\begin{quote}
Find every natural number $m \ge 1$ such that for any real number $x \ge m$, the inequality
\[\sum_{k=1}^m \sqrt{x-k} < x\]
is satisfied.
\end{quote}

Table~\ref{tab:app-ex-math} lists the top ten candidates by embedding cosine similarity. The rank-1 item is character-identical to the query with the inequality reversed; the gold item sits at rank four.

\begin{table}[ht]
\centering
\small
\caption{Top-10 retrieval for query \texttt{sau\_2012\_bb6608}, embedding cosine similarity.}
\label{tab:app-ex-math}
\begin{tabular}{llll}
\toprule
Rank & Cosine & Candidate & Note \\
\midrule
1 & 0.8599 & \texttt{sau\_2012\_bb6608::nm::0} & Wrong top-1: identical text, $<$ flipped to $>$ \\
2 & 0.8396 & \texttt{sau\_2012\_bb6608::nm::2} & Sum replaced with product \\
3 & 0.8311 & \texttt{sau\_2012\_bb6608::eq::medium} & Sibling reformulation \\
4 & 0.8208 & \texttt{sau\_2012\_bb6608::eq::easy} & \textbf{Gold}, ranked fourth \\
5 & 0.8157 & \texttt{sau\_2012\_bb6608::nm::1} & Square roots changed to cube roots \\
6 to 10 & 0.75 to 0.79 & various other-base items & Unrelated problems, lower similarity \\
\bottomrule
\end{tabular}
\end{table}

\subsection{Trajectories: a sibling miss}
\label{app:ex-traj}

Query \texttt{easy\_9} (EASY tier, task type \emph{pick and place simple}, target object \texttt{remotecontrol}, goal receptacle \texttt{sofa}):

\begin{quote}
Place a remote control on the sofa
\end{quote}

GLM-j's chosen candidate, \texttt{alfworld\_286}, is a sibling miss:

\begin{quote}
put a remotecontrol in armchair.
\end{quote}

Same object, same task type, different receptacle: a literal same-object duplicate. This is the sibling failure mode exactly. The judge picked the trajectory that shares the query's literal object instead of one requiring the same procedure over a different object, which is what strict scoring rewards. A representative strict-gold trajectory for this query, \texttt{alfworld\_118}, is:

\begin{quote}
put some newspaper on sofa.
\end{quote}

Different object, same task type, same receptacle as the query: the structural match the judge did not pick.

\section{Per-cell failure taxonomies}
\label{app:taxonomy}

\subsection{Mathematics: baseline retrieval, dominant miss categories}

Table~\ref{tab:app-tax-math} gives the raw counts behind the failure-structure statement in Section 4: the number of strict misses whose rank-1 item is the query's own planted near-miss, and the number whose rank-1 item is a sibling equivalent reformulation.

\begin{table}[ht]
\centering
\caption{Mathematics baseline, rank-1 miss categories by embedder and tier (counts, n=500).}
\label{tab:app-tax-math}
\begin{tabular}{llrr}
\toprule
Embedder & Tier & Own near-miss & Sibling variant \\
\midrule
Gemini-emb & Easy & 420 & 13 \\
Gemini-emb & Hard & 420 & 74 \\
Qwen-emb & Easy & 446 & 7 \\
Qwen-emb & Hard & 446 & 50 \\
\bottomrule
\end{tabular}
\end{table}

\subsection{Trajectories: LLM-reranker miss taxonomy, all nine cells}

Table~\ref{tab:app-tax-traj} gives the per-cell breakdown behind the sibling-dominance statement in Section 5. Percentages are of strict misses within each judge-by-embedder cell.

\begin{table}[ht]
\centering
\small
\caption{Trajectory reranker strict-miss taxonomy, n=118 queries per cell.}
\label{tab:app-tax-traj}
\begin{tabular}{llrrrrr}
\toprule
Embedder & Judge & Misses & Sibling & Near-miss & Other & Unparsed \\
\midrule
Qwen-emb & Gemini-j & 74 & 85.1\% & 13.5\% & 1.4\% & 0.0\% \\
Qwen-emb & GLM-j & 60 & 88.3\% & 11.7\% & 0.0\% & 0.0\% \\
Qwen-emb & Haiku-j & 72 & 84.7\% & 11.1\% & 2.8\% & 1.4\% \\
Gemini-emb & Gemini-j & 70 & 85.7\% & 14.3\% & 0.0\% & 0.0\% \\
Gemini-emb & GLM-j & 50 & 80.0\% & 14.0\% & 2.0\% & 4.0\% \\
Gemini-emb & Haiku-j & 71 & 85.9\% & 12.7\% & 1.4\% & 0.0\% \\
MiniLM & Gemini-j & 74 & 74.3\% & 20.3\% & 5.4\% & 0.0\% \\
MiniLM & GLM-j & 65 & 72.3\% & 7.7\% & 9.2\% & 10.8\% \\
MiniLM & Haiku-j & 80 & 75.0\% & 17.5\% & 5.0\% & 2.5\% \\
\bottomrule
\end{tabular}
\end{table}

\section{Utility grader prompt}
\label{app:grader}

A single grading prompt is built once per query and condition and passed to both graders (Grader A, Gemini 3.1 Flash-Lite; Grader B, GLM-5.2). \texttt{\{problem\}}, \texttt{\{reference\}}, and \texttt{\{candidate\}} are substituted with the query text, the concatenated reference solution(s), and the solver's candidate solution before sending.

\begin{lstlisting}
You are grading a candidate solution to a mathematics olympiad problem against a reference solution.

PROBLEM:
{problem}

REFERENCE SOLUTION:
{reference}

CANDIDATE SOLUTION:
{candidate}

Score the candidate solution's correctness on a scale from 0 to 7, where 7 means fully correct (or containing only minor errors that don't affect the core reasoning) and 0 means completely incorrect or no meaningful progress. Judge whether the candidate's mathematical reasoning and final conclusion are consistent with the reference, not writing style or presentation. The reference may be in a different language than the candidate -- grade the mathematical content, not the language.

Respond with ONLY the integer score (0-7). No explanation, no other text.
\end{lstlisting}

\end{document}